# INTERPRETING DNN OUTPUT LAYER ACTIVATIONS: A STRATEGY TO COPE WITH UNSEEN DATA IN SPEECH RECOGNITION

*Vikramjit Mitra[*], Horacio Franco*

University of Maryland, College Park, MD, USA
Speech Technology and Research Laboratory, SRI International, Menlo Park, CA, USA
`vmitra@umd.edu, horacio.franco@sri.com`

## ABSTRACT

Unseen data can degrade performance of deep neural net (DNN) acoustic models. To cope with unseen data, adaptation techniques are deployed. For unlabeled unseen data, one must generate some hypothesis given an existing model, which is used as the label for model adaptation. However, assessing the goodness of the hypothesis can be difficult, and an erroneous hypothesis can lead to poorly trained models. In such cases, a strategy to select data having reliable hypothesis can ensure better model adaptation. This work proposes a data-selection strategy for DNN model adaptation, where DNN output layer activations are used to ascertain the goodness of a generated hypothesis. In a DNN acoustic model, the output layer activations are used to generate target class probabilities. Under unseen data conditions, the difference between the most probable target and the next most probable target is decreased compared to the same for seen data, indicating that the model may be uncertain while generating its hypothesis. This work proposes a strategy to assess a model's performance by analyzing the output layer activations by using a distance measure between the most likely target and the next most likely target, which is used for data selection for performing unsupervised adaptation.

## 1. INTRODUCTION

Deep learning technologies have revolutionized automatic speech recognition (ASR) systems [1, 2], demonstrating impressive performance for almost all tried languages. Interestingly, deep neural network (DNN)-based systems are both data hungry and data sensitive [3], where the performance of a model is found to improve with additional diverse training data. Unfortunately, annotated training data can be expensive. Although large volumes of data are becoming available every day, not all of it is properly transcribed or reflective of the varying acoustic conditions that systems are expected to tackle. In limited data conditions, DNN acoustic models can be quite sensitive to acoustic-condition mismatches, where subtle variation in the background acoustic conditions can significantly degrade recognition performance.

To cope with the problem of unseen data, multi-condition training accompanied by data augmentation is generally used to expose the DNN acoustic model to a wider range of background acoustic variations [4]. Data augmentation may expose the model to the anticipated acoustic variations; but in reality, acoustic variations are difficult to anticipate. Real-world ASR applications encounter diverse acoustic conditions, which are mostly unique and hence difficult to anticipate. Systems that are trained with several thousands of hours of data collected from different realistic conditions typically are found to be quite robust to background conditions, as they are expected to contain many variations; however, such data may not contain all the possible variations found in the world.

Recently, several open speech recognition evaluations [5-8] have shown how vulnerable DNN acoustic models are to realistic, varying, and unseen acoustic conditions. One of the most celebrated and least resource-constrained approaches to coping with unseen data conditions is performing unsupervised adaptation, where the only necessity is having raw data. A more reliable adaptation technique is supervised adaptation, which assumes having annotated target-domain data; however, annotated data is often unavailable in real-world scenarios. This constraint often makes unsupervised adaptation more practical.

Unsupervised speaker adaptation of DNNs has been explored in [8–11], with adaptation based on maximum likelihood linear regression (MLLR) transforms [10], i-vectors [11], etc. showing impressive performance gains over un-adapted models. In [12] Kullback-Leibler divergence (KLD) based regularization was proposed for DNN model parameter adaptation. Feature-space MLLR (fMLLR) transform was found to improve DNN acoustic model performance for mismatched cases in [13]. Confidence score based unsupervised adaptation demonstrated improvements in recognition performance for Wall Street Journal (WSJ) [14] and VERBMOBIL [15] speech recognition tasks. A semi-supervised DNN acoustic model training was investigated in [16], where a DNN trained with a small dataset was adapted to a larger data set, leveraging data selection using a confidence measure.

In this work, we focus on understanding how acoustic-condition mismatch between the training and the testing data impacts the DNN output decision. Similar efforts have been pursued by researchers in [17, 18]. Earlier [19], we investigated an entropy measure to ascertain the level of uncertainty in a DNN and to translate that measure to quantify DNN decision reliability. This paper focuses on how data mismatch impacts the output layer activations of a DNN, and proposes a measure that predicts when a DNN's decision may be less accurate. The proposed approach relies on the fact that under seen conditions, the most likely

---

target's probability is substantially higher than the next most likely target's probability, whereas for unseen conditions, the difference between those target probabilities may not be as large, which happens as a consequence of the DNN being more uncertain while making a decision in the unseen condition. A similar observation about the impact of unseen data on the winning neuron's activation with respect to the next best activation was cited in [20]. In this work, we use the output layer neural activations (before nonlinear transform) to compute a distance measure between the most likely target and the 2$^{nd}$ and 3$^{rd}$ most likely targets, respectively. We name this measure the confusion distance (*CD*) and show that the *CD* is higher for seen data as compared to unseen data. We compute an averaged distance measure over an utterance and use that to select data for unsupervised adaptation. Note that the proposed strategy is not only restricted to speech recognition but can be used in other applications that involve probabilistic processing.

## 2. DATA

The acoustic models in this work were trained by using the multi-conditioned, noise- and channel-degraded training data from the 16 kHz Aurora-4 [21] noisy WSJ0 corpus. Aurora-4 contains a total of six additive noise types (car; babble; restaurant; street; airport; and train station), with channel-matched and mismatched conditions. It was created from the standard 5K WSJ0 database and contains 7K training utterances of approximately 15-hours duration and 330 test utterances. The test data includes 14 test sets from two different channel conditions and six different added noises. The signal-to-noise ratio (SNR) for the test sets varied between 0-15 dB. Audio data for test sets 1–7 was recorded with a Sennheiser microphone, while test sets 8–14 were recorded using a second microphone randomly selected from a set of 18 different microphones. Results from the test sets are presented as follows: Set A: clean, matched-channel (test set 1); Set B: noisy, matched-channel (test sets 2–7); Set C: clean, varying-channels (test set 8); and Set D: noisy, varying-channels (test sets 9–14).

We treated reverberation as the unseen data condition in our experiments, where we trained the models using the Aurora-4 corpus and evaluated their performance on real-world reverberated data. For *adaptation*, *optimization*, and *evaluation* purposes, we used the *training, development*, and *evaluation* sets distributed with the REVERB-2014 challenge. The REVERB-2014 dataset [8] contains single-speaker utterances, where only the single-microphone part of the dataset was used in the experiments reported in this paper. The REVERB-2014 training set consists of the clean WSJCAM0 [22] data, which was convolved with room impulse responses (with reverberation times from 0.1 sec to 0.8 sec) and then corrupted with background noise. Note that as the REVERB-2104 training set was used as the unsupervised adaptation set, its transcriptions were not used in any of our experiments. The evaluation and development data contain both real recordings (real data) and simulated data (sim data). The real data is borrowed from the MC-WSJ-AV corpus [23], which consists of utterances recorded in a noisy and reverberant room. The simulated evaluation set contains 1088 utterances in each of the far- and near-microphone conditions, and the real evaluation set contains 372 utterances split equally between far- and near-microphone conditions.

We used gammatone filterbank energies (GFBs) as the acoustic features for our experiments. GFBs were generated using a bank of 40 gammatone filters equally spaced on the equivalent rectangular bandwidth scale. The analysis window was 26 ms with a frame rate of 10 ms. The gammatone subband powers were dynamic-range compressed using 15$^{th}$ root. GFBs were used in our experiment because of their robustness against background distortions compared to mel-scale features [27].

## 3. THE CONFUSION DISTANCE (CD) MEASURE

In the case of unknown acoustic variations, DNN-based acoustic models fail to generalize well and, as a consequence, propagate any distortion in the input feature space, resulting in distorted outputs that do not represent relevant aspects of the input [17, 18]. In grossly mismatched situations, detecting the cases that cause the system to completely fail versus those that generate a reasonable output is quite useful. One way to generate such detection is through a confidence measure, which is generally indicative of how trustworthy the ASR hypothesis is for each of the test files. A fully connected network can be interpreted as a cascade of several feature-transformation steps, where the goal is making each target class as discriminative as possible with respect to each other. Hence, for cases where the model fails to generate reasonable performance, such transformations fail to generate reliable features, and therefore the model decision is impacted. It can be expected that when the model decision is impaired, that is when the model is uncertain about its decision, and thus multiple output activations may be generating similar posterior probabilities. A natural indicator of this is how close the neural net activation producing the maximum value is with respect to the activations producing the second or third maxima, respectively. In the case when the distance between the most likely target (i.e., the activation producing the maximum value) and the next most likely target (the activation producing the second-highest maximum value) is less, then the model can be expected to be uncertain while making a decision relative to a model with a greater distance between these values. Note that this distance measure is not an absolute measure, and in this work, we pose it as a relative measure, based on the distances obtained from the training set. We name this distance measure the confusion distance (*CD*), where figure 1 shows its distribution from a training set and an unseen dataset.

Let us assume that a DNN has *N* target classes, indicating *N* neurons in the output layer each generating activations $x_{t,i}$ at a given instant of time *t*, for $i^{th}$ neuron in the output layer. Let $X_t$ be the vector of $x_{t,i}$ at time instant *t*. Let us define $Y_t$ to be the vector obtained after sorting (in descending order) $X_t$. Let the elements of $Y_t$ be $y_{t,i}$, where

$$y_{t,1} \geq y_{t,2} \geq y_{t,3} \geq \cdots \geq y_{t,N}$$

Where

$$y_{t,1} = max_i[x_{t,1}, x_{t,2}, x_{t,3}, \ldots x_{t,N}] = max_i X_t \quad (1)$$

Let us define the frame level CD measure as

$$CD_i = \frac{1}{\alpha}\sum_{i=1}^{\alpha} y_{t,i} - \frac{1}{\beta}\sum_{j=\alpha+1}^{\alpha+\beta} y_{t,j} \quad (2)$$

where the first term determines the average of the top $\alpha$ hypothesis and the second term determines the average of the top $\beta$ competing hypothesis at time instant *t*. For an utterance consisting of *m* frames, the overall averaged CD measure ($CD_{avg}$) is computed by taking the mean of the $CD_i$ estimated from all the m frames.

$$CD_{avg} = \frac{1}{m}\sum_{i=1}^{m} CD_i \quad (3)$$

In this work, the $CD_{avg}$ is estimated for each file for both the training set and the unseen dataset (in this case the unsupervised adaptation set). Let the $CD_{avg}$ computed from the training and the unsupervised adaptation set be denoted as: $CD_{avg}^{train}$ and $CD_{avg}^{unsup\_adapt}$ respectively. The data selection from the unsupervised adaptation set is performed by thresholding the $CD_{avg}^{unsup\_adapt}$ of that set, where the threshold is determined by the $CD_{avg}^{train}$.

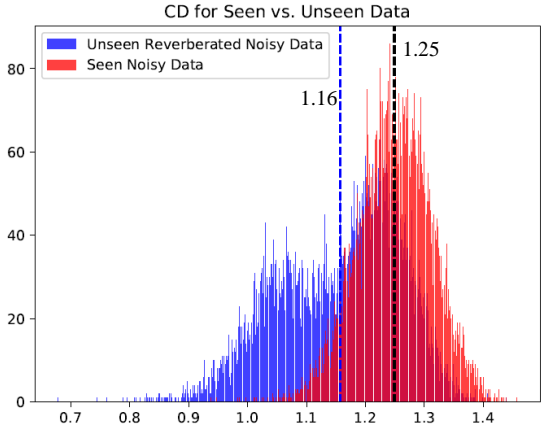

Figure 1. Distribution of CD estimated from the output layer of a DNN acoustic model. Green: CD estimated from the training data (seen noisy). Blue: CD estimated from the unseen data (unseen reverberated + noisy). The vertical dotted lines indicate the respective CD means.

## 4. ACOUSTIC MODEL

In this work, we used time-frequency CNN (TFCNN) acoustic models based on their reliable performance [25, 26] on the Aurora-4 speech recognition task. To generate the alignments necessary for training the acoustic model, a Gaussian mixture model (GMM)-HMM model was used to produce senone labels. Altogether, the GMM-HMM system produced 3125 context-dependent states for the Aurora-4 training data. The input features to the acoustic models were formed by using a context window of 17 frames (8 frames on either side of the current frame). The acoustic models were trained by using cross-entropy on the alignments from the GMM-HMM system. A 5-layered DNN with 2048 neurons in each layer was trained by using the alignments from the GMM-HMM system, which in turn was used to generate alignments for training the subsequent TFCNN acoustic model used in this paper. For the TFCNN acoustic models, the input acoustic features were formed by using a context window of 17 frames. The TFCNNs had 75 filters to perform time convolution and 200 filters to perform frequency convolution. For time and frequency convolution, eight bands were used, followed by a max-pooling over five and three samples, respectively. Feature maps after both the convolution operations were concatenated and fed to a fully connected 4-hidden layer neural net, containing 2048 neurons.

## 5. RESULTS

The baseline acoustic model (TFCNN$_{BASELINE}$) was trained with the Aurora-4 multi-condition training dataset, where a held-out, cross-validation set was used to train the TFCNN acoustic models. The reverberated acoustic condition was treated as the unseen data condition in our experiments, where the experimental analysis was performed by using the development and test data from the REVERB-2014 challenge dataset. As a baseline unsupervised adapted system (TFCNN$_{UV}$) (where the subscript UV stands for unsupervised adaptation), we used the hypothesis from the whole adaptation set to adapt the TFCNN$_{BASELINE}$ model. Note that during adaptation, the unsupervised adaptation dataset was used in addition to the original Aurora-4 training dataset to update the acoustic model parameters. During adaptation, all model parameters were updated with an L$_2$ norm of 0.001 and an initial learning rate of 0.004, with the learning rate halved at every iteration over the adaptation set. Early stopping was performed based on the cross-validation error. Tables 1-3 show the word error rate (WER) obtained from the baseline model (TFCNN$_{BASELINE}$) and unsupervised adapted baseline model (TFCNN$_{UV}$).

Table 1. WERs from the baseline acoustic models when evaluated on the Aurora-4 test set.

| System | Aurora-4 | | | | |
|---|---|---|---|---|---|
| | A | B | C | D | Avg. |
| TFCNN$_{BASELINE}$ | 2.9 | 5.7 | 5.6 | 14.3 | 9.2 |
| TFCNN$_{UV}$ | 3.2 | 5.9 | 6.2 | 14.7 | 9.5 |

Table 2. WERs from the baseline acoustic models when evaluated on the REVERB-2014 dev set.

| System | REVERB 2014 dev | |
|---|---|---|
| | Avg. Sim | Avg. Real |
| TFCNN$_{BASELINE}$ | 39.3 | 42.4 |
| TFCNN$_{UV}$ | 24.4 | 33.7 |

Table 3. WERs from the baseline acoustic models when evaluated on the REVERB-2014 test set.

| System | REVERB 2014 test | |
|---|---|---|
| | Avg. Sim | Avg. Real |
| TFCNN$_{BASELINE}$ | 37.8 | 46.9 |
| TFCNN$_{UV}$ | 22.7 | 37.4 |

Table 1 show that while some performance degradation occurred under the noisy condition with which the model was initially trained, but the degradation is not substantial, which is a consequence of adding the original training set as part of the adaptation set. Tables 2-3 show that using the entire adaptation set improved the model's performance on the unseen reverberation condition for both the dev and test sets of REVERB-2014, reducing the WER by more than 20%. At this point, the question remains if the adaptation step has suffered from any inaccurate hypothesis generated from the adaptation data. Such inaccurate hypothesis can be filtered out by performing data selection by using $CD_{avg}^{unsup\_adapt}$. As an initial experiment to assess the values of $\alpha$ and $\beta$ (refer to equation 2), we rank-sorted the $CD_{avg}^{unsup\_adapt}$ values estimated for the adaptation set and selected the top 4K files for performing adaptation, results shown in table 4. The adapted model after data selection is represented as TFCNN$_{UV\_DS}$, where the subscript DS stands for data selection.

Table 4. WERs from the adapted acoustic models, (where the adaptation set was selected based on different values of $\alpha$ and $\beta$) when tested on the REVERB-2014 dev set.

| System | $\alpha$ | $\beta$ | REVERB-2014 dev | |
|---|---|---|---|---|
| | | | Avg. Sim | Avg. Real |
| TFCNN$_{UV\_DS}$ | 1 | 1 | 28.4 | 35.1 |
| TFCNN$_{UV\_DS}$ | 1 | 2 | 22.3 | 31.5 |
| TFCNN$_{UV\_DS}$ | 1 | 3 | 22.6 | 32.7 |
| TFCNN$_{UV\_DS}$ | 1 | 4 | 22.5 | 32.5 |
| TFCNN$_{UV\_DS}$ | 1 | 5 | 22.5 | 32.9 |
| TFCNN$_{UV\_DS}$ | 2 | 3 | 22.5 | 32.4 |
| TFCNN$_{UV\_DS}$ | 3 | 3 | 28.5 | 36.5 |

Table 4 shows that selecting $\alpha = 1$ and $\beta = 2$ gave the best adaptation performance, at which point the TFCNN$_{UV\_DS}$ acoustic model outperforms the TFCNN$_{UV}$ model. Finally, we explored data selection by using a *CD* threshold learned from the training list. Let $\mu_{train\_CD}$ and $\sigma^2_{train\_CD}$ be the mean and variance computed from the $CD_{avg}^{train}$ data. We can select data using a threshold $\theta$, where the data having $CD_{avg}^{unsup\_adapt} > \theta$ will be selected for performing the unsupervised model adaptation. Table 5 presents the WER results after adaptation, using the data selected with different thresholds, when evaluated on the REVERB-2014 dev set. Table 5 indicates that the optimal threshold was $\theta = \mu_{train\_CD} - 2\sigma_{train\_CD}$. Tables 6-7 present the WERs from the baseline models and the adapted models (using $\alpha = 1$, $\beta = 2$ and $\theta = \mu_{train\_CD} - 2\sigma_{train\_CD}$) for the Aurora-4 and REVERB 2014 eval. sets.

Table 5. WERs from the adapted acoustic models, with data selection using different values of $\theta$ (using $\alpha = 1$ and $\beta = 2$), when evaluated on the REVERB-2014 dev set.

| System | $\theta$ | REVERB-2014 dev | |
|---|---|---|---|
| | | Avg. Sim | Avg. Real |
| TFCNN$_{UV\_DS}$ | $\mu_{train\_CD}$ | 25.3 | 34.0 |
| TFCNN$_{UV\_DS}$ | $\mu_{train\_CD} - \sigma_{train\_CD}$ | 23.1 | 32.6 |
| TFCNN$_{UV\_DS}$ | $\mu_{train\_CD} - 2\sigma_{train\_CD}$ | 22.3 | 31.2 |
| TFCNN$_{UV\_DS}$ | $\mu_{train\_CD} - 3\sigma_{train\_CD}$ | 23.3 | 32.4 |
| TFCNN$_{UV\_DS}$ | $\mu_{train\_CD} - 4\sigma_{train\_CD}$ | 23.6 | 33.8 |

Table 6. WERs from the baseline systems and TFCNN$_{UV}$ (after data selection) acoustic models when evaluated on the Aurora-4 test set.

| System | Aurora-4 | | | | |
|---|---|---|---|---|---|
| | A | B | C | D | Avg. |
| TFCNN$_{BASELINE}$ | 2.9 | 5.7 | 5.6 | 14.3 | 9.2 |
| TFCNN$_{UV}$ | 3.2 | 5.9 | 6.2 | 14.7 | 9.5 |
| TFCNN$_{UV\_DS}$ | 3.3 | 6.0 | 6.1 | 14.7 | 9.5 |

Table 6 shows no substantial change in performance on the seen data (Aurora-4 test set) from the adapted models compared to the TFCNN$_{BASELINE}$ system.

Table 7. WERs from the baseline systems and TFCNN$_{ADAPTED}$ (after data selection) acoustic models when evaluated on the REVERB-2014 test set.

| System | REVERB 2014 test | |
|---|---|---|
| | Avg. Sim | Avg. Real |
| TFCNN$_{BASELINE}$ | 37.8 | 46.9 |
| TFCNN$_{UV}$ | 22.7 | 37.4 |
| TFCNN$_{UV\_DS}$ | 21.1 | 35.0 |

Table 7 shows that data selection followed by model adaptation resulted in better performance than using the entire adaptation data, where a 7% and 6% relative reduction in WER was respectively obtained from data selection compared to using the whole data. This indicates that the data-selection process helps to filter out some bad hypotheses from being used during adaptation. Table 7 shows that the unsupervised adaptation using data selection helped to reduce the WER, where the relative WER improvement were 44% and 25% for simulated and real test data., respectively. The substantial improvement on the simulated reverberation condition is to some extent expected, as the adaptation set used in this case was the REVERB-2014 training set, which consists of simulated reverberation only; hence, it helped the model to learn that condition more than the real reverberation condition.

## 6. CONCLUSION

In this work, we investigated using output layer activations to predict the reliability of a neural net's decision and then using that information to perform data selection for unsupervised model adaptation. We proposed a metric, the confusion distance (*CD*), and used it to perform data selection for performing unsupervised adaptation. A lower *CD* reflects more confusion in a neural net hypothesis stemming from the reduced distance between the winning target and the next most probable target. We used data that resulted in higher *CD* values for doing model adaptation and demonstrated that filtering out data with bad hypotheses resulted in relative WER improvement of 6 to 7%. In this work, we used a summary *CD* measure ($CD_{avg}$) for each utterance; however, the measure can also be obtained at the individual frame level, providing frame-level confusion information. Future studies should explore using a frame-level confidence measure while selecting data segments for performing unsupervised adaptation.

## 7. ACKNOWLEDGEMENTS

This material is based upon work partly supported by the Defense Advanced Research Projects Agency (DARPA) under Contract No. HR0011-15-C-0037. The views, opinions, and/or findings contained in this article are those of the authors and should not be interpreted as representing the official views or policies of the Department of Defense or the U.S. Government.